\documentclass[acmtog]{acmart}

\acmSubmissionID{311}

\usepackage{booktabs} 

\usepackage[ruled]{algorithm2e} 

\SetAlFnt{\small}
\SetAlCapFnt{\small}
\SetAlCapNameFnt{\small}
\SetAlCapHSkip{0pt}

\acmJournal{TOG}

\begin{document}
\title{One-shot Compositional 3D Head Avatars with Deformable Hair}

\author{Yuan Sun}
\affiliation{%
  \institution{Xi'an Jiaotong University}
  \city{Xi'an}
  \country{China}
}
\author{Xuan Wang}
\affiliation{%
  \institution{Xi'an Jiaotong University}
  \city{Xi'an}
  \country{China}
}
\author{WeiLi Zhang}
\affiliation{%
  \institution{Xi'an Jiaotong University}
  \city{Xi'an}
  \country{China}
}
\author{Wenxuan Zhang}
\affiliation{%
  \institution{Xi'an Jiaotong University}
  \city{Xi'an}
  \country{China}
}
\author{Yu Guo}
\affiliation{%
  \institution{Xi'an Jiaotong University}
  \city{Xi'an}
  \country{China}
}
\author{Fei Wang}
\affiliation{%
  \institution{Xi'an Jiaotong University}
  \city{Xi'an}
  \country{China}
}

\begin{abstract}
We propose a compositional method for constructing a complete 3D head avatar from a single image. Prior one-shot holistic approaches frequently fail to produce realistic hair dynamics during animation, largely due to inadequate decoupling of hair from the facial region, resulting in entangled geometry and unnatural deformations. Our method explicitly decouples hair from the face, modeling these components using distinct deformation paradigms while integrating them into a unified rendering pipeline. Furthermore, by leveraging image-to-3D lifting techniques, we preserve fine-grained textures from the input image to the greatest extent possible, effectively mitigating the common issue of high-frequency information loss in generalized  models.
Specifically, given a frontal portrait image, we first perform hair removal to obtain a bald image. Both the original image and the bald image are then lifted to dense, detail-rich 3D Gaussian Splatting (3DGS) representations. For the bald 3DGS, we rig it to a FLAME mesh via non-rigid registration with a prior model, enabling natural deformation that follows the mesh triangles during animation. For the hair component, we employ semantic label supervision combined with a boundary-aware reassignment strategy to extract a clean and isolated set of hair Gaussians.
To control hair deformation, we introduce a cage structure that supports Position-Based Dynamics (PBD) simulation, allowing realistic and physically plausible transformations of the hair Gaussian primitives under head motion, gravity, and inertial effects. Striking qualitative results, including dynamic animations under diverse head motions, gravity effects, and expressions, showcase substantially more realistic  hair behavior alongside faithfully preserved facial details, outperforming state-of-the-art one-shot  methods in perceptual realism. 
\end{abstract}




\begin{teaserfigure}
  \centering
  \includegraphics[width=1.\linewidth]{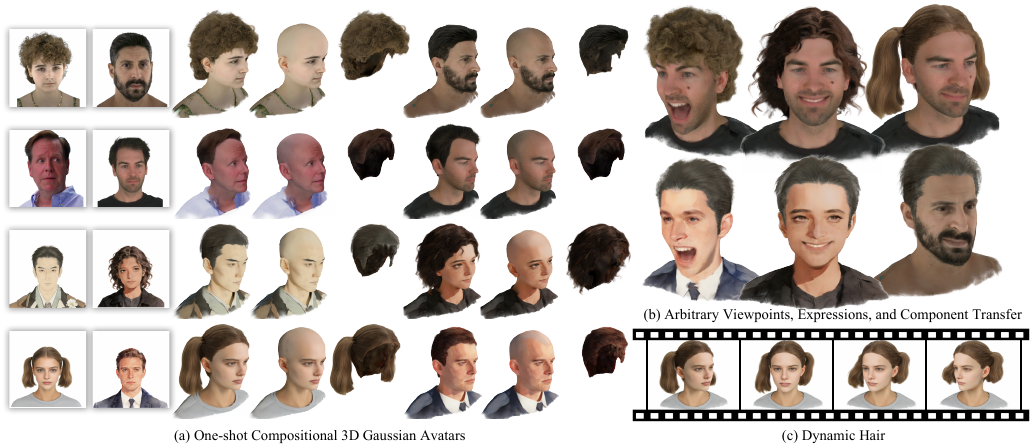}
  \caption{We introduce a novel method that reconstructs decoupled 3D Gaussian head avatars from a single input image. These avatars support effortless hairstyle transfer and enable real-time rendering from arbitrary viewpoints and facial expressions, featuring natural dynamic hair motion.}
  \Description{}
  \label{teaser}
\end{teaserfigure}

\maketitle

\section{Introduction}
High-fidelity reconstruction of animatable 3D head avatars is crucial for immersive applications such as virtual try-on, video games, and virtual reality, where both photorealistic appearance and physically plausible dynamics remain key objectives.
Most existing methods \cite{beeler2011high,teotia2024hq3davatar,fyffe2014driving,ma2021pixel,saito2024relightable,qian2024gaussianavatars,wang20253d,giebenhain2024npga,xu2024gaussian,lee2024surfhead,cai2025hera,aneja2025scaffoldavatar,li2025tega} rely heavily on large-scale, person-specific multi-view video data captured in controlled lab settings with calibrated cameras \cite{gafni2021dynamic,kirschstein2023nersemble,pan2023renderme,wuu2022multiface,martinez2024codec}, severely restricting accessibility for real-world deployment. Recent advances have explored efficient reconstruction under minimal data regimes, leveraging strong priors from large-scale pre-training to regress expressive 3D representations directly from one or few casual images \cite{kirschstein2025flexavatar,wu2026uika,kirschstein2025avat3r,zhao2024invertavatar,khakhulin2022realistic,li2023generalizable,cha2025perse,teotia2024gaussianheads,teotia2024hq3davatar,teotia2025audio}. However, these prior-based approaches often trade off person-specific fidelity, struggling to recover subtle geometric and textural details unique to individuals. More critically, their holistic joint modeling of face and hair as a single entity disregards the inherent decomposability of the human head, resulting in unrealistic hair dynamics and appearance artifacts during animation and novel-view synthesis.

While several recent works have pursued compositional head modeling to mitigate these issues, significant practical limitations persist: many still depend on multi-view sequences \cite{liao2025hhavatar,wang2025mega,wang2023neuwigs}, others construct isolated high-quality hair models without seamless integration into a complete, real-time-capable full-head pipeline \cite{luo2024gaussianhair,sklyarova2025im2haircut}, and even explicit face–hair separation methods frequently neglect or superficially handle realistic hair dynamics \cite{kim2025haircup,he20253dgh}.

To overcome these challenges, we propose a method that reconstructs a fully compositional, controllable, and animatable 3D head avatar from a single input image, with explicit disentanglement of face and hair components to enable physically plausible hair deformation under diverse head motions and expressions. To the best of our knowledge, we are the first to achieve this setting.  Specifically, given a frontal portrait, we first employ an image editing model to generate a bald version of the input. Both the original and bald images are then lifted to static 3D Gaussian Splatting (3DGS) \cite{kerbl20233d} representations using FaceLift \cite{lyu2025facelift}. For the bald representation, we leverage a rigging information prior and perform joint reconstruction and registration to rig it onto the FLAME parametric mesh \cite{li2017learning}, while completing missing internal structures such as teeth and eyeballs. This registration strategy preserves the high-fidelity texture details and strong identity preservation obtained from the lifting stage. For the haired representation, we utilize semantic segmentation from multi-view images sampled at arbitrary viewpoints as supervision to isolate the hair region, followed by a boundary-aware  reassignment strategy for refinement. For long-haired identities, we construct a coarse cage and apply Mean Value Coordinates (MVC) \cite{ju2023mean} to drive Gaussian deformation during cage transformations, with physically plausible dynamics governed by Position-Based Dynamics (PBD) simulation \cite{muller2007position,macklin2016xpbd}. Using proxy-based collision constraints to achieve plausible collision simulation.

In summary, our contributions are as follows: We present the first one-shot method for reconstructing a fully compositional and animatable 3D head avatar from a single frontal image, in which the face and hair components are driven by distinct deformation paradigms and seamlessly integrated into a unified differentiable rendering pipeline, addressing a key gap in prior single-image approaches. We further propose a purely simulation-driven cage-based deformation framework for physically plausible hair dynamics, without relying on any data-driven learning, and introduce a novel proxy-based collision constraint that ensures realistic and stable collision simulation.

\section{Related work}

\subsection{One-shot Animatable Head Avatar}
Exploiting priors from large-scale datasets, one-shot methods can regress complete 3D head avatars using textured meshes \cite{yang2020facescape,khakhulin2022realistic,zielonka2022towards}, feature grids \cite{deng2024portrait4d,ki2024learning,sun2023next3d,li2023generalizable,ma2023otavatar,tran2024voodoo,liu2025avatarartist}, or NeRFs \cite{zhuang2022mofanerf,rebain2022lolnerf,yu2023nofa}, often conditioned on 3DMM \cite{blanz2023morphable,li2017learning} expression latents or learned deformation fields. Despite enabling novel-view synthesis and expression control, these approaches still struggle with fine-grained details and real-time performance.

The recent rise of 3D Gaussian Splatting (3DGS) has brought substantial improvements in both fidelity and rendering speed. Generative methods such as GAIA \cite{yu2025gaia} and AGORA \cite{fazylov2025agora} leverage conditional GANs or adversarial training for photorealistic, controllable avatars. Prior-based methods, such as HeadGAP \cite{zheng2025headgap} and GASP \cite{saunders2025gasp}, learn generalizable Gaussian priors from large-scale or synthetic datasets, enabling high-fidelity personalization. Diffusion-augmented approaches, including CAP4D \cite{taubner2025cap4d} and related works \cite{yin2025facecraft4d,taubner2025mvp4d}, enrich training via Stable Diffusion–based augmentation before per-subject fitting. GAGAvatar \cite{chu2024gagavatar} introduces a novel dual-lifting method that achieves high fidelity while bypassing the slow per-identity optimization or multi-stage codec pipelines prevalent in prior methods. Fast feed-forward models such as LAM \cite{he2025lam} further enable rapid, generalizable reconstruction. Collectively, these advances have pushed the boundaries of rendering quality, controllability, efficiency, and cross-identity generalization.

Nevertheless, a common limitation remains: most methods entangle facial and hair geometry without explicit disentanglement. This coupling frequently causes physically implausible hair behaviors under large-amplitude head motions or rapid pose changes. In contrast, our proposed method delivers vivid facial animation while ensuring hair deforms in a physically plausible manner under distinct motion dynamics.

\subsection{3D Dynamic Hair Modeling}
Early hair modeling in academia and industry predominantly relied on simplified geometric primitives (e.g., 2D strips \cite{koh2000real,liang2003enhanced,noble2004modelling}, wisps \cite{chen1999system,choe2005statistical,yang2000cluster,xu2001v,patrick2004modelling},  multi-resolution cylinders \cite{wang2004hair,kim2002interactive}, explicit meshes \cite{yuksel2009hair}). While efficient for rendering and basic animation, these representations struggled to reproduce complex hair geometry and realistic strand-level dynamics.
This limitation motivated the shift toward explicit strand-based representations \cite{piuze2011generalized,xing2019hairbrush,shen2023ct2hair,bitterlipractical,fascione2018path,sklyarova2023haar}, which better capture individual fiber behavior and support physically-based simulation \cite{fei2017multi,daviet2023interactive,hsu2023sag,zhou2025augmented}. 

With the emergence of 3D Gaussian Splatting (3DGS), several works explored Gaussian-based hair representations. Methods like GaussianHair \cite{luo2024gaussianhair} and Gaussian Haircut \cite{zakharov2024human} introduced specialized Gaussian primitives, achieving impressive static reconstruction quality. However, they typically treat hair as static geometry and delegate animation to traditional graphics engines, making it challenging to achieve fully joint and consistent deformation of hair, head, and facial expressions.
HADES \cite{liao2025hades} addressed this by proposing a unified, strand-based 3DGS framework that enables end-to-end differentiable rendering of hair together with the head and face. Yet strand-based approaches remain computationally heavy.
An alternative line of recent methods exploits 3DGS’s ability to represent high-frequency volumetric details and applies lightweight deformation fields to directly manipulate hair Gaussians at runtime \cite{liao2025hhavatar,wang2025mega}. These approaches offer excellent efficiency and real-time performance, but often rely on multi-view video capture for training and show limited generalization to novel identities or large head poses. We also note several recent works that reconstruct hair from a single image \cite{sklyarova2025im2haircut,wang2025srm}; however, these methods primarily focus on static hair reconstruction and do not jointly render hair together with the face.

We propose a simple yet effective cage-based deformation control scheme for hair manipulation. During inference, the cage vertices are directly deformed through Position-Based Dynamics (PBD) \cite{muller2007position} simulation, enabling physically plausible motion without requiring multi-view temporal data for training. The deformation of Gaussian primitives is governed by Mean Value Coordinate (MVC) \cite{ju2023mean} from the cage, ensuring smooth and stable propagation of motion. As a result, our method achieves real-time, vivid, and temporally stable rendering.

\begin{figure*}[t]
    \centering
    \includegraphics[width=1\linewidth]{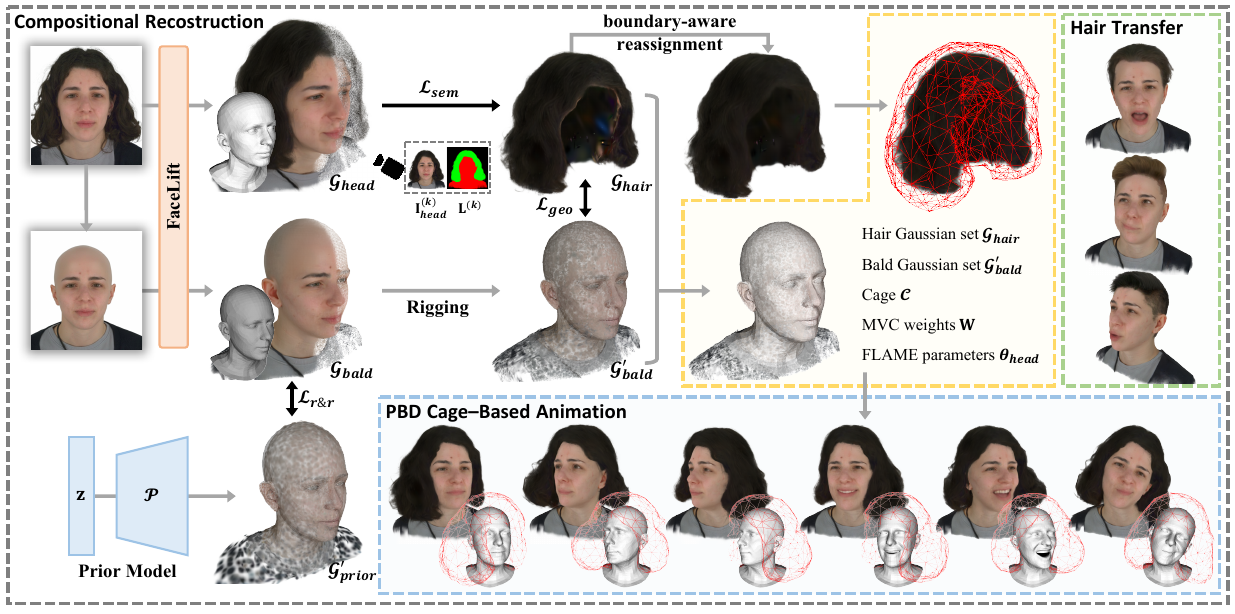}
    \vspace{-18pt}
    \caption{\textbf{Method Overview.} Given a single frontal image, we explicitly decouple hair and bald face components for separate reconstruction using 3DGS. The bald part is lifted to 3DGS and rigged to a parametric FLAME mesh via non-rigid registration for natural expression-driven deformation. The hair Gaussians are isolated and enclosed in a cage structure that supports Position-Based Dynamics (PBD) simulation for physically plausible motion under head pose, gravity, and inertia. Both components are seamlessly integrated into a unified rendering and animation pipeline, enabling coherent and realistic full-head driving. The driving stage procedure is briefly summarized in Algorithm \ref{alg:inference_t_to_t+1}.}
    \label{fig:framework}
    \vspace{-15pt}
\end{figure*}
\section{Method}

\subsection{Preliminary}
\textbf{Rigged 3D Gaussian Splatting.}
3D Gaussian Splatting (3DGS)  \cite{kerbl20233d} represents scenes as a set of 3D Gaussian ellipsoids $\mathcal{G} = \{\mathscr{g}_n\}^N_{n=1}$, where each Gaussian $\mathscr{g} = \{\boldsymbol{\mu}, \mathbf{r}, \mathbf{s}, \alpha, \mathbf{c}\}$ defines position, rotation, scale, opacity, and color. The rendering is performed by differentiable rasterization: $\mathbf{I} = \mathcal{R}_{\mathbf{c}}(\mathcal{G}, \mathbf{T})$, with $\mathbf{T}$ denoting the pose of the camera. Convention: subscript on $  \mathcal{R}  $ indicates the the primary attribute being rendered ($  \mathcal{R}_\mathbf{c}  $ for color, $  \mathcal{R}_d  $ for depth, $  \mathcal{R}_\alpha  $ for opacity,  etc.). To achieve controllable dynamic expressions, GaussianAvatars (GA)  \cite{qian2024gaussianavatars} employs a rigging   strategy that anchors 3D Gaussian primitives to a deformable FLAME mesh  \cite{li2017learning}. Each Gaussian primitive is defined in the local coordinate system of its parent mesh triangle, indicated by the binding index $b$, and represented as  $\mathscr{g}' = \{\boldsymbol{\mu}', \mathbf{r}', \mathbf{s}', \alpha, \mathbf{c}, b\}$. When the mesh deforms, these local attributes are converted to world coordinates using a rigid affine transform computed from the host triangle:
\begin{equation}
\mathbf{r}  = \mathbf{R} \mathbf{r}^{\prime}, 
\boldsymbol{\mu}  = \eta \mathbf{R} \boldsymbol{\mu}^{\prime} + \mathbf{t}, 
\mathbf{s} = \eta \mathbf{s}^{\prime}.
\end{equation}
where $\mathbf{t}$, $\mathbf{R}$, and $\eta$ are derived from the centroid, orientation, and deformation scale of the triangle. Given a FLAME mesh $\mathcal{M}$ parameterized by $ \boldsymbol{\theta} $, the transformation of a local Gaussian primitive $\mathscr{g}^{\prime}$ to the global space is expressed as:
$
\mathscr{g} = \mathcal{T}_{l2g}(\mathscr{g}^{\prime}, \mathcal{M}(\boldsymbol{\theta})).
$
The inverse transformation is given by:  
$
\mathscr{g}^{\prime} = \mathcal{T}_{g2l}(\mathscr{g}, \mathcal{M}(\boldsymbol{\theta}),b).
$

\noindent \textbf{Cage-based Deformation.}
Mean Value Coordinates (MVC) allow robust deformation of points inside a coarse control cage~\cite{ju2023mean}.
For a source cage $\mathcal{C}_s$ with vertices $\{\mathbf{c}_{s,j}\}^J_{j=1}$, any interior point
$\mathbf{x}_s \in \mathbb{R}^3$ is expressed as a convex combination using fixed MVC weights $w_j$:
\begin{equation}
\label{mvc}
\mathbf{x}_s = \sum_j w_j \mathbf{c}_{s,j} .
\end{equation}
These weights $w_j$ are computed once on the source cage.
When the cage is deformed to $\mathcal{C}_d$ with vertices $\mathbf{c}_{d,j}$,
the deformed point is obtained by reusing the same weights:
\begin{equation}
\label{mvc1}
\mathbf{x}_d = \sum_j w_j \mathbf{c}_{d,j} .
\end{equation}
This produces a smooth, continuous deformation field, enabling intuitive editing of enclosed detailed geometry via direct cage manipulation.
This formulation has recently been extended to 3DGS deformation~\cite{huang2024gsdeformer,tong2025cage,xie2024sketch}.

\noindent \textbf{Position-Based Dynamics.} PBD  achieves real-time physical simulation by iteratively enforcing geometric constraints directly on vertex positions  \cite{macklin2014unified,muller2007position}. Extended Position-Based Dynamics (XPBD) further improves this framework by enabling compliance parameters that control stiffness independently of solver iterations and time steps  \cite{macklin2016xpbd}. PBD and its extensions excel in interactive applications, such as video games and virtual reality, providing robust handling of collisions, attachments, and complex deformations.

\subsection{Compositional Head Recostruction}
Our purpose is reconstructing a component-wise 3D head representation from a single image by decomposing it into two 3D Gaussian primitives sets: a hair component $\mathcal{G}_{hair}$ and a hairless head component $\mathcal{G}_{bald}$. The two components are animated under distinct deformation paradigms, enabling independent control of facial motion and hair dynamics.
Given a frontal portrait, we first apply a 2D bald filter  \cite{wu2022hairmapper,zhang2025stable} to transform it into a hairless image . We then apply FaceLift  \cite{lyu2025facelift} to both the original input image and the edited bald image to reconstruct high-fidelity static 3D Gaussian representations, yielding $\mathcal{G}_{head}$ and $\mathcal{G}_{bald}$, respectively. The  image-to-3D lifting strategy provides rich geometric and texture details. As a natural byproduct of this process, we can also readily recover the corresponding  FLAME meshes, denoted as $\mathcal{M}(\boldsymbol{\theta}_{head})$ and $\mathcal{M}(\boldsymbol{\theta}_{bald})$. To obtain multi-view supervision for subsequent optimization, we uniformly sample $  K  $ viewpoints on the sphere and render the ground-truth images from the initial Gaussians as follows: $  \{\mathbf{I}_{head}^{(k)}\}_{k=1}^{K} = \{\mathcal{R}_{\mathbf{c}}(\mathcal{G}_{head}, \mathbf{T}^{(k)})\}_{k=1}^{K} $ and  $  \{\mathbf{I}_{bald}^{(k)}\}_{k=1}^{K} = \{\mathcal{R}_{\mathbf{c}}(\mathcal{G}_{bald}, \mathbf{T}^{(k)})\}_{k=1}^{K} $.

\noindent \textbf{Hair Part Segmentation.} Following a similar strategy to several recent segmentation approaches in 3D Gaussian Splatting~ \cite{zhou2024feature,wu2024opengaussian}, we augment each 3D Gaussian primitive with a learnable low-dimensional attribute $  \mathbf{f} \in \mathbb{R}^2  $.
We obtain the corresponding hair/non-hair semantic labels $  \{ \mathbf{L}^{(k)} \}_{k=1}^{K}  $ by applying SAM2~ \cite{ravi2024sam2} to the multi-view head images $  \{ \mathbf{I}_{\text{head}}^{(k)} \}_{k=1}^{K}  $. These binary labels serve as supervision for the rasterized per-pixel feature maps $  \mathbf{F}^{(k)} = \mathcal{R}_{\mathbf{f}} ( \mathcal{G}_{\text{head}}, \mathbf{T}^{(k)} )  $, obtained via differentiable feature rendering:
\begin{equation}
\mathcal{L}_{sem}=\mathcal{L}_1(\mathbf{F}^{(k)},\mathbf{L}^{(k)}).
\end{equation}
After only a modest number of optimization iterations, the hair Gaussians $  \mathcal{G}_{\text{hair}}  $ are readily disentangled via a softmax operation applied to the learned feature $  \mathbf{f}  $.

\noindent \textbf{Rigging Bald Part by Registration.} For the static $  \mathcal{G}_{bald}  $, once each Gaussian primitive is assigned a reasonable corresponding rigging information, it can naturally deform in accordance with the FLAME mesh. Following the approach of  \cite{zheng2025headgap,sun2025fine}, we leverage a prior model $  \mathcal{P}  $ pre-trained on a large-scale dataset to provide this information. Specifically, the prior model can obtain a set of Gaussian primitives defined in the mesh-local coordinate system, $\mathcal{G}_{prior}' = \mathcal{P}(\mathbf{z})$, by decoding an identity latent code $\mathbf{z}$. Given $\mathbf{C}_{prior}^{(k)}=\mathcal{R}_{\mathbf{c}}(\mathcal{T}_{l2g}(\mathcal{G}_{prior}', \mathcal{M}(\boldsymbol{\theta}_{bald})), \mathbf{T}^{(k)})$ and $\mathbf{C}_{bald}^{(k)}=\mathcal{R}_{\mathbf{c}}(\mathcal{G}_{bald}, \mathbf{T}^{(k)})$, we optimize the identity latent code $  \mathbf{z}  $, $ \mathcal{P} $, $\boldsymbol{\theta}_{bald}$ , and $  \mathcal{G}_{bald}$ using a joint reconstruction and registration loss:
\begin{equation}
\begin{aligned}
    \mathcal{L}_{r\&r}&= \mathcal{L}_{rec}(\mathbf{C}_{prior}^{(k)},\mathbf{I}_{bald}^{(k)}) + \mathcal{L}_{rec}(\mathbf{C}_{bald}^{(k)},\mathbf{I}_{bald}^{(k)}) \\&+\lambda_1\mathcal{L}_{chamfer}(\{\boldsymbol{\mu}_{prior}\},\{\boldsymbol{\mu}_{bald}\}),
\end{aligned}
\end{equation}
where $\mathcal{L}_{rec}=(1-\lambda_2)\mathcal{L}_{1} + \lambda_2\mathcal{L}_{ssim} + \lambda_3\mathcal{L}_{lpips}$.  The Chamfer distance term \cite{li2022non} $  \mathcal{L}_{chamfer}  $ encourages structural similarity between the two Gaussian point sets . 
After non-rigid registration, each Gaussian primitives $  \mathscr{g}_{bald} $ in global space can transfer into triangle-local coordinates by
$  \mathscr{g}_{bald}' = \mathcal{T}_{g2l}(\mathscr{g}_{bald}, \mathcal{M}(\boldsymbol{\theta}_{bald}),b^*) $,
where $  b^* $ is acquired from its nearest neighbor $  \mathscr{g}_{prior}' $. We replaced the teeth and eyeball regions in $\mathcal{G}_{bald}'$ with the corresponding parts from $\mathcal{G}_{prior}'$, determined according to the FLAME faces indices, to obtain a complete  bald head.

\noindent \textbf{Component Assembly.}  Although current 2D bald filter techniques are highly advanced and introduce almost no loss, the results from the dual image-to-3D pipeline do not preserve shape consistency, leading to a noticeable gap between $\mathcal{M}(\boldsymbol{\theta}_{head})$ and $\mathcal{M}(\boldsymbol{\theta}_{bald})$. Fortunately, since we have obtained $\mathcal{G}_{bald}'$ defined in the local coordinate, $\mathcal{G}_{hair}$ and $\mathcal{G}_{bald} = \mathcal{T}_{l2g}(\mathcal{G}_{bald}', \mathcal{M}(\boldsymbol{\theta}_{head}))$ can be easily roughly aligned and combined by transforming the FLAME coefficients. At this stage, the combined result does not guarantee that the scalp and hair regions are free from interpenetration. We therefore fix the parameters of $\mathcal{G}_{bald}'$ and joint optimize $\mathcal{G}_{hair}$  along with the FLAME coefficients $\boldsymbol{\theta}_{head}$ to ensure that the hair lies outside the scalp:
\begin{equation}
\begin{aligned}
     \mathcal{L}_{geo}&= \mathcal{L}_{rec}(\mathbf{C}_{head}^{(k)},\mathbf{I}_{head}^{(k)}) + \lambda_4\mathcal{L}_{sem}(\mathbf{F}_{head}^{(k)},\mathbf{L}^{(k)})\\&+ \lambda_5\mathcal{L}_{collision}(\{\boldsymbol{\mu}_{hair}\},\mathcal{M}(\boldsymbol{\theta}_{head}),\epsilon),
\end{aligned}
\end{equation}
where $\mathbf{C}_{head}^{(k)}=\mathcal{R}_{\mathbf{c}}(\mathcal{G}_{hair} \cup \mathcal{T}_{l2g}(\mathcal{G}_{bald}', \mathcal{M}(\boldsymbol{\theta}_{head})), \mathbf{T}^{(k)})$. The term $  \mathcal{L}_{collision}  $ serves as a collision penalty  \cite{grigorev2023hood,santesteban2022snug,chen2025d} that ensures the hair point set $  \{\boldsymbol{\mu}_{hair}\}  $ stays outside the head mesh $  \mathcal{M}(\boldsymbol{\theta}_{head})  $, allowing a small tolerance margin $  \epsilon  $.

\begin{figure}[thbp]
  \centering
  \includegraphics[width=0.47\textwidth ]{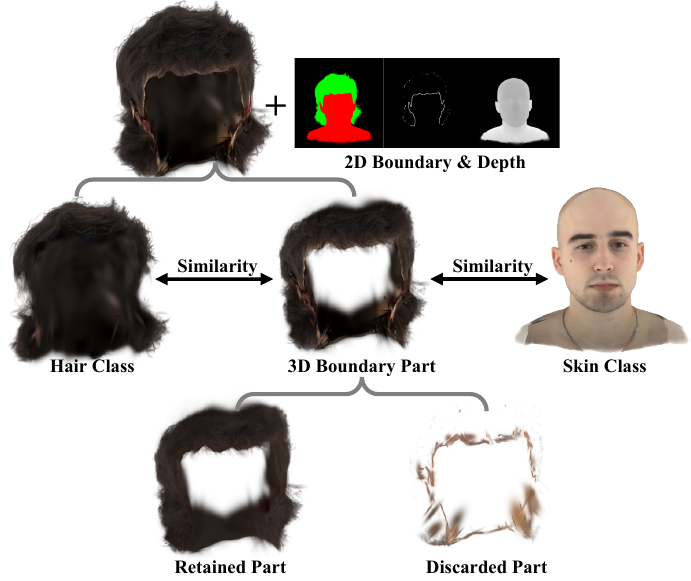}
  \vspace{-10pt}
  \caption{\textbf{Hair Cleanup via Boundary-aware Reassignment.} We extract the 3D boundary region using 2D boundary and depth information. Within the local neighborhood, we measure the similarity of each Gaussian to hair and skin classes, then reassign it accordingly. This effectively eliminates residual skin contamination caused by inaccurate 2D segmentation.}
  \label{fig:boundary}
\end{figure}

Due to pixel-level inaccuracies in the semantic labels, $  \mathcal{G}_{hair}  $ often contains residual skin Gaussians near the boundaries. To address this, we propose a \textbf{boundary-aware reassignment} strategy (Fig. \ref{fig:boundary}). By combining the 2D boundaries from $  \mathbf{L}^{(k)}  $ with depth information in $  \mathcal{G}_{head}  $, we divide $  \mathcal{G}_{hair}  $ into boundary and non-boundary subsets. The non-boundary hair Gaussians, along with $  \mathcal{G}_{bald}  $, serve as two class centers. Boundary Gaussians are then reclassified based on similarity to these centers using per-Gaussian color and scale features, effectively removing misclassified skin primitives and enabling a natural transition between hair and face regions. 

By defining $\mathscr{g}_{hair}$ in the local coordinates of the nearest scalp mesh, rigid transformations with head pose and hairstyle transfer across different identities can be easily achieved. For identities with short hair, this already yields a complete and fully   head avatar $\{\mathcal{G}_{hair}', \mathcal{G}_{bald}', \boldsymbol{\theta}_{head}\}$ which can be driven by modified expression or poses coefficients in $\boldsymbol{\theta}_{head}$. However, for long hair, rigid transformations of the hair are physically inaccurate; we describe the deformation strategy for such cases in the next section.


\subsection{Hair Deformation with Cage-PBD}
Our goal is to achieve real-time hair deformation that seamlessly follows facial expressions. Inspired by GSDeformer \cite{huang2024gsdeformer}, we adopt a coarse cage-based deformation approach. Specifically, we perform physics-based simulation using Position-Based Dynamics (PBD) on a lightweight cage mesh with fewer than 500 vertices. The deformed cage vertices are then used to propagate the deformation to the hair Gaussian primitives via Mean Value Coordinates (MVC) interpolation, as formulated in Eq. \eqref{mvc1}.
By simulating just the lightweight cage instead of all Gaussians directly, and running everything in the high-performance Taichi framework \cite{hu2019taichi,hu2021quantaichi,hu2019difftaichi}, we obtain efficient, real-time hair dynamics while preserving fine details.

\noindent \textbf{Deform Gaussian Primitives with Cage.}  
Given a source static hair Gaussian set $  \mathcal{G}_{hair}=\{\mathscr{g}_{hair}\}^{N}_{n=1}  $, we first follow the methods in \cite{xian2009automatic,xu2022deforming} to construct a watertight source cage $  \mathcal{C}_s  $ by voxelizing the dense point cloud $  \{\boldsymbol{\mu}_{hair}\}^{N}_{n=1}  $. Drawing inspiration from \cite{huang2024gsdeformer}, each Gaussian ellipsoid $\mathscr{g}_{hair}$ is represented  using $ \mathcal{X}_s = \{\mathbf{x}_s^{c},\mathbf{x}_s^{x+},\mathbf{x}_s^{x-},\mathbf{x}_s^{y+},\mathbf{x}_s^{y-},\mathbf{x}_s^{z+},\mathbf{x}_s^{z-}\}  $, which consists of the ellipsoid center and its six axis endpoints.
The mean value coordinates (MVC) weights $  \mathbf{W} \in \mathbb{R}^{N \times 7 \times M}  $ are then computed by solving Eq.~\eqref{mvc}, where $  M  $ is the number of vertices in the cage. For a detailed description of the computational procedure, we refer the reader to the original work \cite{ju2023mean,xu2022deforming}.
Once the cage is deformed into  $  \mathcal{C}_d  $, the deformed positions $ \mathcal{X}_d = \{\mathbf{x}_d^{c},\mathbf{x}_d^{x+},\mathbf{x}_d^{x-},\mathbf{x}_d^{y+},\mathbf{x}_d^{y-},\mathbf{x}_d^{z+},\mathbf{x}_d^{z-}\}  $ can be efficiently obtained using Eq.~\eqref{mvc1}.  Unlike prior work \cite{huang2024gsdeformer} that reconstructs deformed Gaussian primitives via SVD decomposition and splitting, we adopt a lightweight approximation tailored to hair modeling. Since hair Gaussian primitives are typically slender ellipsoids, we only consider the deformation of their principal axis.
Specifically, in the source configuration, the principal axis is identified as
$i^* = \arg\max_{i \in \{x,y,z\}} \left\| \mathbf{x}_s^{i+} - \mathbf{x}_s^{i-} \right\|.$
After deformation, the Gaussian center is computed as the midpoint of the transformed endpoints along the principal axis:
\begin{equation}
\boldsymbol{\mu}_d = \frac{1}{2} \left( \mathbf{x}_d^{i^*+} + \mathbf{x}_d^{i^*-} \right).
\end{equation}
The rotation is updated by applying a relative rotation $\Delta \mathbf{R}$ induced by the principal-axis deformation to the original rotation, while the scale is adjusted according to the corresponding axis-length ratio:
\begin{equation}
   \mathbf{r}_d = \Delta \mathbf{R} \, \mathbf{r}_s, \quad
\mathbf{s}_d = \mathbf{s}_s \cdot \frac{\left\| \mathbf{x}_d^{i^*+} - \mathbf{x}_d^{i^*-} \right\|}{\left\| \mathbf{x}_s^{i^*+} - \mathbf{x}_s^{i^*-} \right\|}.
\end{equation}

\noindent \textbf{Cage Deformation by PBD.}
To achieve physically plausible deformation of the cage mesh while simulating realistic hair dynamics (roots follow scalp motion and drag the tips), we employ a PBD solver~\cite{macklin2014unified,muller2007position,macklin2016xpbd}. Cage vertices are treated as particles. Vertices near the scalp (hair roots) are designated as kinematic particles with zero inverse mass; their positions are directly prescribed by Linear Blend Skinning (LBS) driven by head motion at each time step. In other words, these points deform exactly like the vertices on the FLAME scalp region, following the same LBS weights and joint transformations. Given a time step from $t$ to $t + \Delta t$, our goal is to compute the cage $\mathcal{C}^{(t+\Delta t)}$. Consider the current cage $\mathcal{C}^{(t)}$ with particle positions $\{\mathbf{c}_j^{(t)}\}_{j=1}^{J}$ and velocities $\{\mathbf{v}_j^{(t)}\}_{j=1}^{J}$.
For the free (non-kinematic) vertices, we first perform a semi-implicit Euler integration to obtain predicted positions:
\begin{equation}
\label{pbd_predict}
  \mathbf{v}_j^{(t)} \leftarrow \mathbf{v}_j^{(t)} + \Delta t \cdot \beta_j \cdot \mathbf{g}_{j}^{(t)}, \quad \mathbf{p}_j^{(t)} \leftarrow \mathbf{c}_j^{(t)} + \Delta t \cdot \mathbf{v}_j^{(t)}
\end{equation}
where $\beta_j = 1/m_j$ is the inverse mass, $\mathbf{g}_{j}^{(t)}$ denotes the gravity, $\Delta t$ is the time step, and $\mathbf{p}_j^{(t)}$ is the predicted (unconstrained) position for vertex $j$.
These predicted positions serve as the initial guess for the subsequent constraint projection phase. We iteratively project (typically 15 iterations) the predicted positions onto the constraint manifold to obtain the final positions at the next time step:
\begin{equation}
\label{pbd_constraint}
    \mathbf{c}_j^{(t+\Delta t)} = \Pi_{\text{Constraint}} \bigl( \mathbf{p}_j^{(t)} \bigr) \quad \forall j,
\end{equation}
where $\Pi_{\text{Constraint}}$ denotes the approximate projection operator defined by the set of standard geometric constraints, including stretch (length), bending, volume preservation, and collision handling, following common practices in Position-Based Dynamics (PBD) and its extensions such as XPBD~\cite{muller2007position,macklin2016xpbd,wu2023two,Zhang:CompDynamics:2020}.
After projection, the velocities are updated for the next time step using the finite-difference approximation:
\begin{equation}
\label{pbd_updatev}
    \mathbf{v}_j^{(t+\Delta t)} \leftarrow \frac{\mathbf{c}_j^{(t+\Delta t)} - \mathbf{c}_j^{(t)}}{\Delta t}.
\end{equation}
This formulation naturally reproduces root-dragging behavior: kinematic roots pull free particles through the constraints, yielding stable swing, inertia, and recovery even at large time steps.

\begin{figure}[thbp]
  \centering
  \includegraphics[width=0.47\textwidth ]{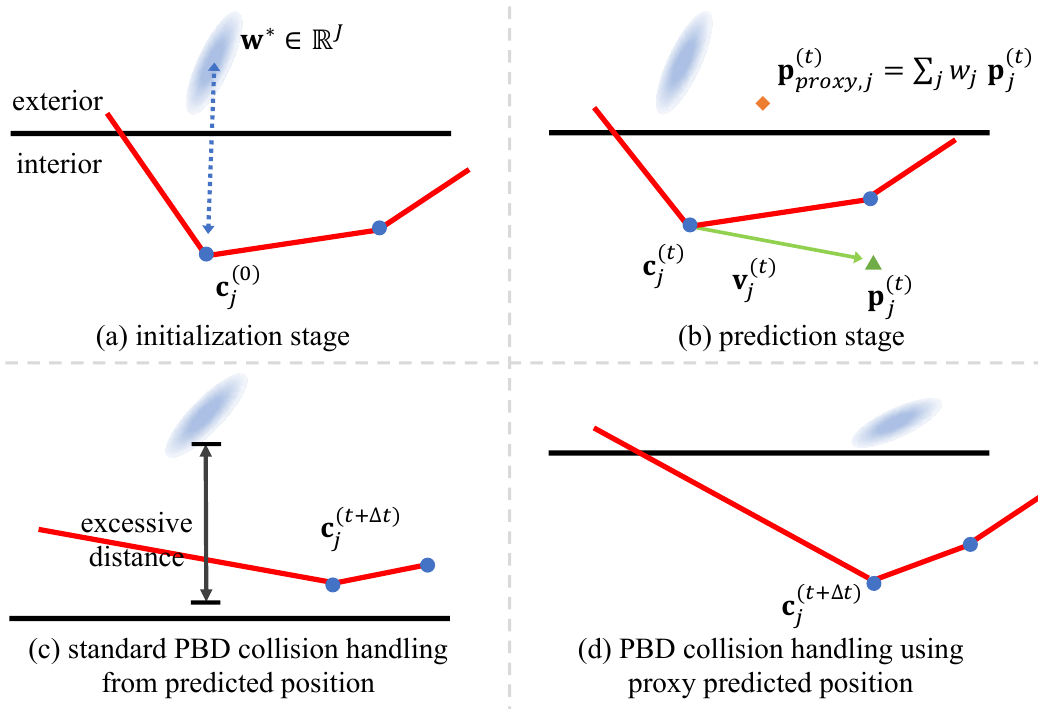}
  \vspace{-10pt}
  \caption{\textbf{Proxy-based collision constraint}. The black solid curve denotes a cross-sectional slice of the FLAME mesh, while the red solid curve indicates the edges of the deformation cage. During cage construction, for each cage vertex particle, we record its MVC weights with respect to the centers of its nearest neighboring Gaussian primitives. During collision detection, instead of directly testing the predicted particle positions $\mathbf{p}^{(t)}_j$, we interpolate a proxy point using the precomputed MVC weights and perform collision detection at this proxy point $\mathbf{p}^{(t)}_{proxy,j}$, while applying the resulting corrections to the predicted particle positions. As illustrated in state (b) of the figure, directly correcting $  \mathbf{p}^{(t)}_j  $ would push the cage vertices away from the mesh surface, resulting in excessively large gaps (c); in contrast, performing collision detection and correction at the proxy point avoids such over-repulsion and maintains reasonable adherence without requiring additional collision constraints (d).}
  \label{fig:pbd_collision}
  \vspace{-10pt}
\end{figure}

A key observation is that the cage we construct is not tightly fitted to the underlying Gaussian point set, inevitably leaving non-negligible gaps between the cage vertices and the Gaussians. This structural characteristic prevents us from directly applying conventional collision constraints between cage vertices and the FLAME mesh surface.
To address this issue, we introduce a \textbf{proxy-based collision constraint} strategy, as illustrated in Fig. \ref{fig:pbd_collision}. During cage initialization, we assign each cage vertex an MVC weight with respect to the Gaussians, determined simply by the index of its nearest-neighbor Gaussian. After position prediction, we compute the corresponding proxy particle  by applying the pre-assigned MVC weights to the predicted positions (see Fig. \ref{fig:pbd_collision}(b)):
\begin{equation}
\label{proxy}
    \mathbf{p}^{(t)}_{proxy,j}=\sum_{j}w_j\mathbf{p}^{(t)}_j.
\end{equation}
This proxy effectively represents the potential location of the Gaussian splat under the current deformation. When collision resolution is performed:
Directly using the predicted  position would push the point outward along the FLAME surface normal (Fig. \ref{fig:pbd_collision}(c)) and cause an excessively large separation between the hair strands and the skin due to the existing gap between the cage and the Gaussians. By contrast, performing collision detection and correction using the proxy particle instead ensures that displacement is applied only when the underlying Gaussian would genuinely penetrate the FLAME mesh. This proxy-based formulation produces significantly more plausible contact behavior and faithfully preserves the intended hair–skin proximity.

\begin{algorithm}[t]
\caption{Inference: Image Rendering at $t{+}1$}
\label{alg:inference_t_to_t+1}

\KwIn{
$\mathcal{G}'_{\text{bald}}$;
$\mathcal{G}_{\text{hair}}$;
$\boldsymbol{\theta}^{(t+1)}$;
$\mathcal{C}^{(t)}$;
$\mathbf{T}$;
$\mathbf{W}$.
}
\KwOut{Rendered image $\mathbf{I}^{(t+1)}$.}

\textbf{1.} Compute the deformed bald part in global space:

\quad $\mathcal{G}^{(t+1)}_{\text{bald}}
=
\mathcal{T}_{l2g}
(
\mathcal{G}'_{\text{bald}},
\mathcal{M}(\boldsymbol{\theta}^{(t+1)}_{\text{head}})
)
$.

\textbf{2.} Calculate zero inverse-mass cage vertices by LBS.

\textbf{3.} PBD simulation for other vertices:

\For{$i = 1$ \KwTo $n_{substeps}$}{
    $\Delta t = 1 / n_{substeps}$;
    
    \textbf{3.1} Predict particles position by Eq.~\eqref{pbd_predict}\;

    \textbf{3.2} Calculate proxy particles by Eq. ~\eqref{proxy};

    \textbf{3.3} Constraint step by Eq.~\eqref{pbd_constraint}\;

    \textbf{3.4} Update velocities by Eq.~\eqref{pbd_updatev}\;
}

\textbf{4.} Get $\mathcal{G}^{(t+1)}_{\text{hair}}$ by $\mathcal{C}^{(t+1)}$ and $\mathbf{W}$.

\textbf{5.} Rendering: $\mathbf{I}^{(t+1)}
=
\mathcal{R}_{\mathbf{c}}
(
\mathcal{G}^{(t+1)}_{\text{bald}}
\cup
\mathcal{G}^{(t+1)}_{\text{hair}},
\mathbf{T}
)$

\Return $\mathbf{I}^{(t+1)}$
\end{algorithm}

\section{Experiments}
\begin{table*}[t]
    \centering
    \caption{Quantitative results on the monocular video dataset.}
    \label{tab:results}
    \vspace{-0.1in}
    \resizebox{\textwidth}{!}{
    \begin{tabular}{l|ccccccc|ccc}
        \toprule
        & \multicolumn{7}{c|}{\textbf{Self Reenactment}} & \multicolumn{3}{c}{\textbf{Cross Reenactment}} \\
        \textbf{Method} 
        & PSNR$\uparrow$ & SSIM$\uparrow$ & LPIPS$\downarrow$ & CSIM$\uparrow$ & AED$\downarrow$ & APD$\downarrow$ & AKD$\downarrow$ 
        & CSIM$\uparrow$ & AED$\downarrow$ & APD$\downarrow$ \\
        \midrule
        VOODOO3D~\cite{tran2024voodoo} 
        & 16.94 & 0.767 & 0.302 & 0.512 & 0.123 & 0.134 & 6.072 
        & 0.548 & 0.198 & 0.177 \\
        
        GPAvatar~\cite{chu2024gpavatar} 
        & 18.54 & 0.810 & 0.242 & 0.602 & 0.091 & 0.094 & 5.372 
        & 0.522 & 0.164 & 0.158 \\
        
        Portrait4D-v2~\cite{deng2024portrait4d} 
        &  19.97&  0.833&  0.145&  0.767&  0.070 &   0.072&  4.959&   0.645&    0.142&    0.137 \\
        
        GAGAvatar~\cite{chu2024gagavatar} 
        & 20.51 & 0.850 & 0.130 & 0.814 & 0.066 & 0.065 & 4.194 
          & 0.764 &   0.130&   0.132  \\
        
        CAP4D~\cite{taubner2025cap4d} 
        & 20.23 & 0.854 & 0.133 & 0.798 & 0.068 & 0.061 & \textbf{4.122} 
        & 0.720 & 0.132 & 0.128 \\
        
        LAM~\cite{he2025lam} 
        & 19.47 & 0.825 & 0.155 & 0.731 & 0.080 & 0.077 & 5.019 
        & 0.628 & 0.146 & 0.139 \\
        \midrule
        \textbf{Ours} 
        &  \textbf{20.79}&   \textbf{0.864}&  \textbf{0.124}&  \textbf{0.824}&   \textbf{0.059}&   \textbf{0.057}&  4.206&  \textbf{0.768} &    \textbf{0.120}&    \textbf{0.125}\\
        \bottomrule
    \end{tabular}
    }
    \vspace{-0.1in}
\end{table*}
\begin{figure*}[!t]
  \centering
  \includegraphics[width=1\textwidth ]{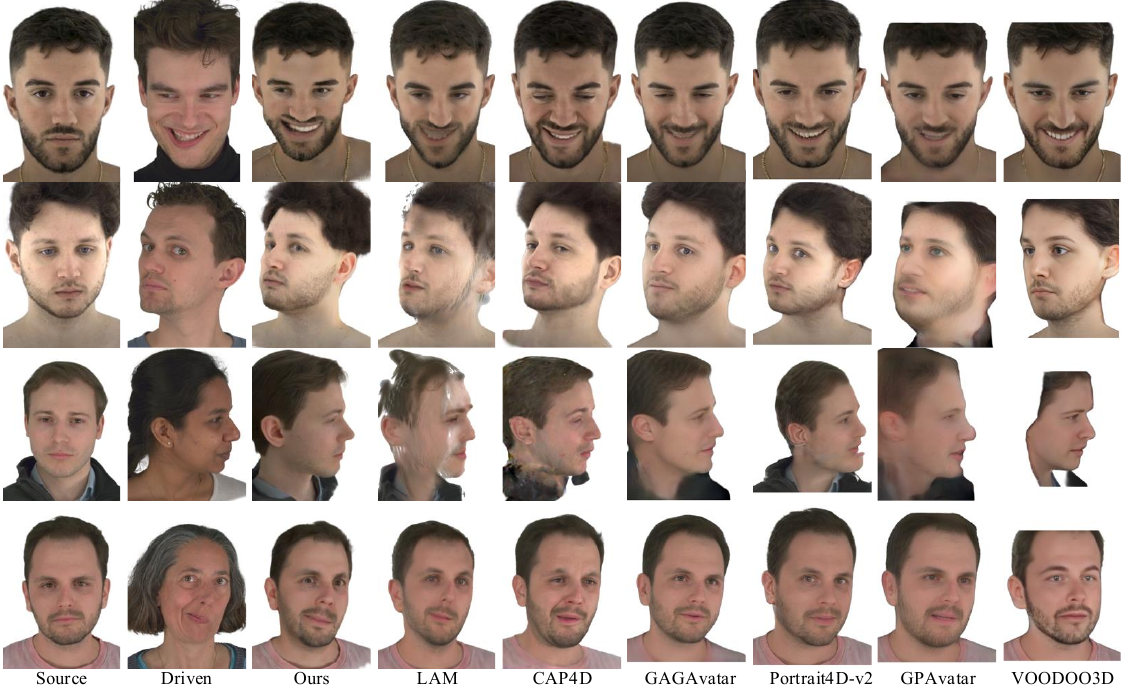}
  \vspace{-20pt}
  \caption{Cross-reenactment comparison.}
  \vspace{-10pt}
  \label{fig:cross-reenactment}
\end{figure*}

\subsection{Experiments Setting}
\textbf{Baselines.} We evaluate our method in comparison with several recent baselines. These include triplane-based approaches (GPAvatar  \cite{chu2024gpavatar}, Portrait4D-v2  \cite{deng2024portrait4d}, and VOODOO3D  \cite{tran2024voodoo}); CAP4D  \cite{taubner2025cap4d}, which leverages a Morphable Multi-View Diffusion Model (MMDM) for training data augmentation; GAGAvatar  \cite{chu2024gagavatar}, based on a dual-lifting strategy; and LAM  \cite{he2025lam}, which uses canonical-space points from FLAME as Transformer queries to predict Gaussian attributes from multi-scale image features.

\noindent \textbf{Datasets.} 
For evaluation, we select 20 frontal sequences each from NeRSemble \cite{kirschstein2023nersemble} and Ava256 \cite{martinez2024codec}, and 10 clean interview clips from VFHQ \cite{xie2022vfhq}, totaling 50 identities covering diverse expressions and head motions.
FLAME parameters for animation driving are extracted using VHAP \cite{qian2024gaussianavatars,qian2024vhap}.
While our method currently struggles with significant external occlusions (e.g., microphones in some VFHQ clips), we exclude such cases from our evaluation. This constitutes a limitation of our approach, which we discuss in more detail in the Limitation section in supplementary material. We believe that this restriction does not significantly affect practical usability in typical real-world scenarios.

\begin{figure*}[thbp]
  \centering
  \includegraphics[width=1\textwidth ]{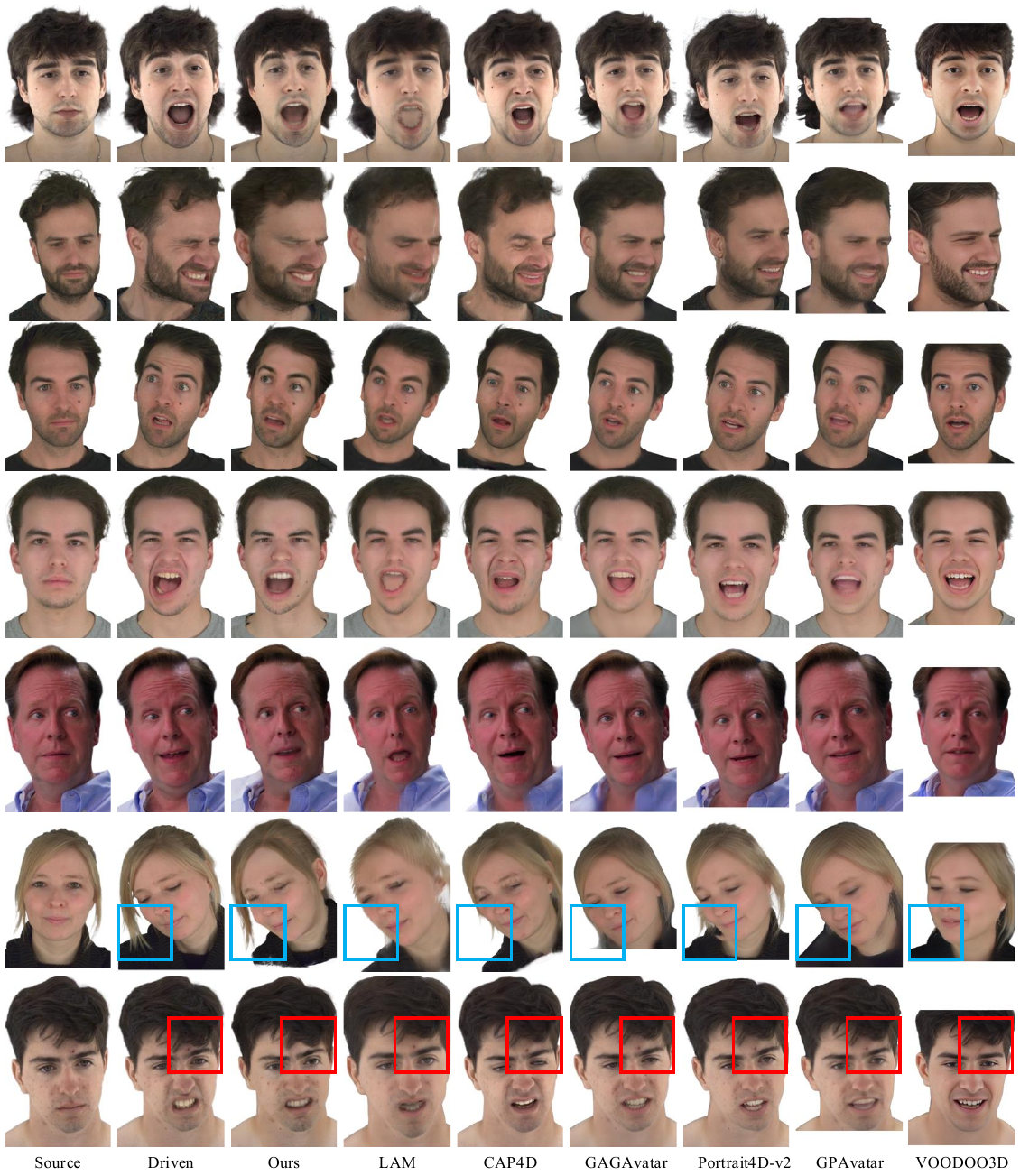}
  \caption{Qualitative comparison on self-reenactment of head avatars.}
  \label{fig:self-reenactment}
\end{figure*}
\noindent \textbf{Metrics.} We use the following commonly used metrics for evaluation. Image quality is assessed with Peak Signal-to-Noise Ratio (PSNR), Structural Similarity Index (SSIM)  \cite{wang2004image}, and Learned Perceptual Image Patch Similarity (LPIPS)  \cite{zhang2018unreasonable}, which measure pixel-level fidelity and perceptual similarity. Identity similarity is evaluated using Cosine Similarity (CSIM) on face recognition features extracted by a pre-trained ArcFace model  \cite{deng2019arcface}. Expression and pose transfer faithfulness are measured by Average Expression Distance (AED) and Average Pose Distance (APD), estimated by the 3DMM-based facial parameter regressor from Deep3DFaceRecon\cite{deng2019accurate}. Geometric accuracy of facial motion is evaluated with Average Keypoint Distance (AKD)  \cite{bulat2017far}, the mean Euclidean distance between corresponding 2D facial landmarks.

\subsection{Experimental analysis}
Table \ref{tab:results} demonstrates that our method delivers superior image quality, excellent identity preservation, and precise expression and pose consistency. By jointly considering the quantitative results and the visual comparisons of self-reenactment and cross-identity reenactment shown in Fig. \ref{fig:self-reenactment} and Fig. \ref{fig:cross-reenactment}, we make the following observations.

Most methods achieve a reasonable approximation of facial expression and head pose reenactment. However, noticeable differences appear in fine-grained detail preservation.
Our approach and GAGAvatar both demonstrate strong fidelity to source image details (e.g., hair and beard texture). This advantage stems from the shared use of direct lifting-based 3D Gaussian representations, which maintain near pixel-level alignment between Gaussians and input pixels. In contrast, encoder–decoder-based methods struggle to faithfully reconstruct such high-frequency details, while CAP4D essentially produces an averaged representation over augmented data, further limiting its ability to capture fine structures. Nevertheless, GAGAvatar relies on dual lifting only for the input view of the source image, and its performance degrades sharply under large head motions or significant viewpoint changes. Additionally, many methods suffer from a lack of volume or poor side-view quality. By relying on a static and complete head representation, our method effectively mitigates this issue.

Another critical difference lies in hair modeling. Other approaches do not explicitly separate hair from the face, leading to unnatural boundary deformations—for example, bangs deform rigidly together with the forehead during eyebrow raising (see red boxes in Fig. \ref{fig:self-reenactment}). In contrast, our method employs PBD-based hair simulation, enabling physically plausible, gravity-driven swaying and dangling of hair strands (see blue boxes in Fig. \ref{fig:self-reenactment}), which is not observed in other methods.
\begin{figure}[thbp]
  \centering
  \includegraphics[width=0.47\textwidth ]{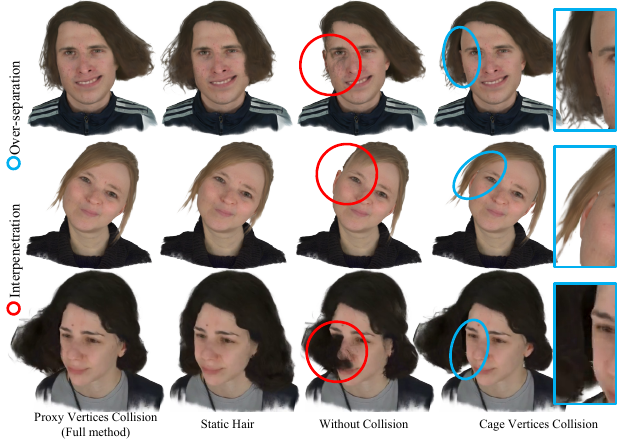}
  \vspace{-5pt}
  \caption{\textbf{Ablation Study on Hair Deformation.} Compared to static hair, PBD-based deformation responds more naturally to head motion. Without collision constraints, interpenetration occurs; applying constraints directly to cage vertices results in large gaps due to imperfect initial alignment between the cage and Gaussians during construction. Our proxy-based method effectively resolves this issue, achieving realistic  hair dynamics.}
  \label{fig:dynamic_hair}
  \vspace{-5pt}
\end{figure}

\subsection{Ablation Study}
\noindent \textbf{Effect of Hair Deformation.}  As evidenced by the  qualitative visualizations in Fig. \ref{fig:dynamic_hair}, static hair without cage-based deformation exhibits unnatural rigidity and fails to produce physically plausible motion during head movements and expressions. Removing our PBD collision constraint causes severe entanglement between face and hair components, resulting in prominent interpenetration artifacts under dynamic animation. Relying solely on cage-vertex collision detection ignores volumetric gaps between Gaussian primitives and cage , leading to  unstable behavior. In contrast, our proxy-based collision constraint, which introduces intermediate proxy points to resolve inter- and intra-component contacts more accurately, yields stable, penetration-free, and intuitively realistic hair dynamics. 

\begin{figure}[h]
  \centering
  \includegraphics[width=0.47\textwidth ]{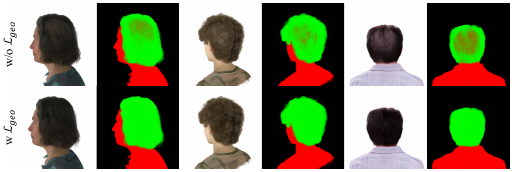}
  \vspace{-10pt}
  \caption{\textbf{Ablation Study on $\mathbf{\mathcal{L}_{geo}}$.} By optimizing the Gaussian primitives of the hair component together with the FLAME parameters, interpenetration at the occipital region can be effectively avoided.}
  \vspace{-5pt}
  \label{fig:flame_shape}
\end{figure}

\noindent \textbf{Effect of Loss $\mathbf{\mathcal{L}_{geo}}$.}
Fig. \ref{fig:flame_shape} shows the rendered color and semantic visualizations, from which we observe that the Gaussian primitives corresponding to hair in $\mathcal{G}_{hair}$ are not guaranteed to lie outside the mesh $\mathcal{M}(\boldsymbol{\theta}_{head})$, particularly in the occipital region. The proposed joint geometric optimization loss $\mathcal{L}_{geo}$ is therefore necessary to enforce geometry consistency.

\begin{figure}[h]
  \centering
  \includegraphics[width=0.47\textwidth ]{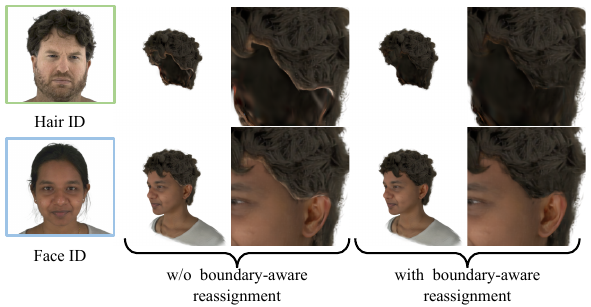}
  \vspace{-10pt}
  \caption{\textbf{Ablation study on boundary-aware reassignment.} This strategy removes residual skin from hair Gaussians, yielding smoother and more natural cross-identity hairstyle transfer.}
  \label{fig:seg_hair}
\end{figure}

\noindent \textbf{Effect of Boundary-aware Reassignment.} As shown in Fig. \ref{fig:seg_hair}, hard-boundary supervision using the segmentation mask is inherently inaccurate, which leads to unnatural sharp artifacts in the boundary regions. In contrast, the proposed boundary-aware reassignment effectively removes residual skin artifacts, enabling smoother transitions and more natural, seamless hairstyle transfer.

\section{Conclusions}

In this work, we present a novel compositional approach for reconstructing animatable 3D full-head avatars from a single frontal portrait image. By explicitly separating the hair and face components, we model each with specialized deformation techniques: rigid FLAME-driven deformation for the face and cage-based Position-Based Dynamics for the hair. This design effectively addresses key challenges in one-shot holistic reconstruction, including entangled geometry, boundary artifacts, and unrealistic hair motion. Future work will explore extensions to unconstrained inputs, strand-level hair dynamics, and physical simulation, aiming to achieve fully controllable, photorealistic 3D head avatars.


\newpage
\bibliographystyle{ACM-Reference-Format}
\bibliography{reference}


\end{document}